\documentclass{ecai}
\usepackage{times}
\usepackage{graphicx}
\usepackage{latexsym}

\usepackage[utf8]{inputenc}
\usepackage[english]{babel}
\usepackage{amsthm}
\usepackage{graphicx}
\usepackage{xcolor}
\usepackage{bbm}
\usepackage{mathtools}
\usepackage{enumitem}
\usepackage{hyperref}
\graphicspath{ {images/} }
\theoremstyle{definition}
\newtheorem{definition}{Definition}[section]
\usepackage{subcaption}
\usepackage{wrapfig}
\usepackage[super]{nth}
\usepackage{wasysym}
\usepackage{multirow}

\newcount\Comments  % 0 suppresses notes to selves in text
\newcommand{\kibitz}[2]{\ifnum\Comments=1{\color{#1}{#2}}\fi}
\newcommand{\ym}[1]{\kibitz{blue}{[YM:#1]}}
\newcommand{\kg}[1]{\kibitz{red}{[KG:#1]}}

\ecaisubmission

\begin{document}

\title{Personalization in Human-AI Teams: Improving the Compatibility-Accuracy Tradeoff}

\author{Jonathan Martinez\institute{Ben-Gurion University, Israel, email: martijon@post.bgu.ac.il} \and Kobi  Gal\institute{Ben-Gurion University, Israel, email: kobig@bgu.ac.il} \and Ece Kamar\institute{Microsoft Research, USA} \and Levi H. S.  Lelis\institute{Amii, University of Alberta, Canada, email: levi.lelis@ualberta.ca} }

\maketitle
\bibliographystyle{ecai}
\begin{abstract}
 AI systems that model and interact with users can update their models over time to reflect new information and changes in the environment. Although these updates may improve the overall performance of the AI system, they may actually hurt the performance with respect to individual users. Prior work has studied the trade-off between improving the system's accuracy following an update and the compatibility of the updated system with prior user experience. The more the model is forced to be compatible with a prior version, the higher loss in accuracy it will incur. In this paper, we  show that by 
 personalizing the loss function to specific users, 
 %in some cases it is possible to  increase the overall prediction accuracy of the AI system without compromising its compatibility \ym{ Maybe this sentence is incorrect? Since the tradeoff is only improved, not solved, meaning that compatibility is compromised to a certain degree}.
 %
%  in some cases it is possible to improve the compatibility-accuracy tradeoff (increase the accuracy of the model  while sacrificing less compatability with respect to the history of the specific  user).
  in some cases it is possible to improve the compatibility-accuracy trade-off with respect to these users (increase the compatibility of the model  while sacrificing less accuracy).
 %
 %improve this compatibility-accuracy trade-off relative to specific users
 %by employing novel loss functions  that personalize the trained model to be compatible with the user's history of interaction with the system. 
 We present experimental results indicating that 
 %on average our method provides only marginal improvements over previous methods, but
 this approach provides moderate improvements on average (around 20\%) but large improvements for certain users (up to 300\%). 
%  %
% In human-AI teams where the AI's goal is to assist the human in the decision-making process, the human's trust in the AI's predictions plays a key role in the team's performance. The AI system may eventually be updated with the goal of increasing its overall accuracy. When the update is not supervised it may achieve this goal while unfortunately introducing mistakes that didn't occur in the pre-update version of the system, potentially violating the mental model of the system that the user developed during his/her interaction with the pre-update version of the system and damaging the user's trust. This kind of update is said to possess low compatibility. Previous work explored methods for inducing varying amounts of compatibility to an update and showed the existence of a compatibility-accuracy
% trade-off. We expand upon these methods by considering ways of increasing the compatibility for a specific user or group of users, i.e., personalizing the updates to improve the compatibility-accuracy trade-off and present experimental results that indicate good improvement when averaging over many users and great improvement for certain users.
\end{abstract}
\section{INTRODUCTION}
\label{sec:intro}
Advancements in AI and ML have led to advice provisioning systems that derive insights and make predictions from large amounts of data. For example, expert diagnostic systems in healthcare predict patients' health condition by analyzing lifestyle, physical health records and social activities, and make suggestions to doctors about possible treatments~\cite{sahoo2019deepreco}.  
% can assist users in the decision-making process in such a way that the performance of the human-AI team is better than the performance of either the AI system or the human alone \cite{kamar2016directions}. This is possible due to humans and AI systems having a different set of capabilities that can be complementary to each other \cite{bansal2019beyond}.
%\kg{can u include a real example e.g. doctor?} For example, the AI system can offer a speed-up in the decision making process by showing the user useful information in a short period of time thanks to its computing capabilities and ability to statistically process large quantities of data while humans can identify the validity of the system's predictions by using common sense or complex knowledge of the problem’s domain that is hard to emulate in a computer.
%
%\par In these human-AI teams, trust plays a key role. It was observed in previous work that the user's lack of trust in the predictions of the system has a negative impact on the performance of the human-AI team \cite{bansal2019updates}. 
%During the user's interaction with a system, the user generates a mental model of it and expects it to behave according to this mental model \cite{norman1988psychology}. This includes complex systems like AI systems \cite{kulesza2012tell}. 
As the user interacts with the system, two processes occur. First, the user develops a mental model of the system's capabilities based on the quality of the recommendations. Second, the system collects more data and is able to update its prediction models. While updating the system model can improve accuracy, it can also change the way the system makes predictions in a way that does not agree with the user's expectations or mental model of the system. Thus while the update improves the overall system performance, it may exhibit a poor compatibility with the user's expectations~\cite{bansal2019updates}, possibly causing the user to lose trust in the system and ignore its recommendations.
%
% The error boundary of the system may change in such a way that the system's behavior starts violating the mental model that the user previously developed and therefore decreases the performance of the team, even though the system's overall accuracy may have increased due to the update. An update that changes the system's error boundary in such a way is referred to as an update with poor 
\par As an example, imagine a doctor that is being assisted by an AI-system that predicts whether skin moles are cancerous or not. Suppose that the system's average accuracy is currently 70\% overall, and that the doctor's speciality is face skin moles. 
%including skin moles located in the patient's arms, the kind of skin moles that this particular doctor deals with the most. 
Next, the system  receives an update which increases its average accuracy to 90\% overall but decreases it to 60\% for face skin moles. As a consequence of this poorly compatible update, the doctor may notice the drop in accuracy regarding this specific region of data and start mistrusting the predictions of the system, therefore missing out on the benefits of being assisted by the the AI-system.
Or worse still, the doctor may not notice this drop and therefore continue trusting the system's predictions that happen to be less accurate than before the update.
\par Bansal et. al. \cite{bansal2019updates} suggested a method for adjusting the compatibility of updates to AI systems where the loss function is modified to incur an additional penalty for \textit{new mistakes}, i.e., mistakes that the system's post-update version makes that the pre-update version didn't make. In the same research they show that a trade-off exists between the compatibility of the update and the accuracy of the updated version of the system: The more an update is forced to be compatible, the less accurate it will be. The method explored by Bansal et al. is designed to lower the amount of new errors relative to the whole data-set, meaning that all instances in the data-set are given the same importance regardless of the user that interacts with the updated system.
\begin{figure}[t]
     \centering
     \includegraphics[width=0.45\textwidth]{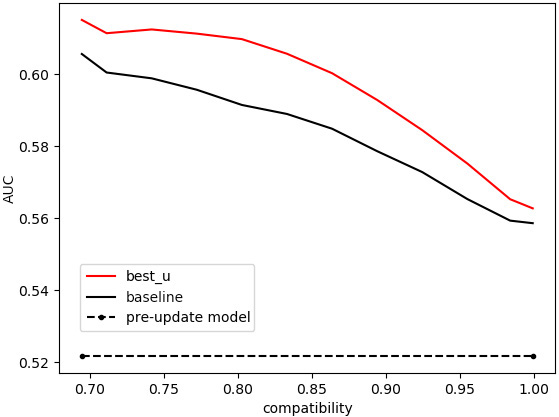}
     \captionsetup{width=.9\linewidth}
     \caption{ Compatibility-accuracy trade-off averaged over all users from the experiment with the edNet data-set. The x and y axes represent the compatibility and AUC of the predictions, respectively. The average trade-off produced by the baseline can be compared with the one produced by our method.}
    \label{fig:example_tradeoff}
\end{figure}
 \par Our hypothesis is that improvements in the compatibility-accuracy trade-off can be achieved by personalizing the update to the AI-system towards specific users. In other words, that increasing the penalty given for wrong predictions on instances that the user usually interacts with can produce better compatibility-accuracy trade-offs (achieve the same degree of compatibility while sacrificing less accuracy) for that user  than previous methods \cite{bansal2019updates}. We propose a method that personalizes the update to the needs of a specific user (Section~\ref{sec:personalization}), a novel metric for measuring the performance of these trade-offs (Section~\ref{sec:methodology}) and then present a series of experiments where our method is shown to perform better than a baseline model that doesn't perform any personalization (Section~\ref{sec:results}). We then analyze two factors that affect the magnitude of the improvements provided by our method – the length of the user's history of interaction with the AI-system and the degree to which the feature distribution of this history differs from the distribution of the whole data-set. The analysis indicates that the correlation between the performance of our method and these two factors varies according to the classification task, but that our method is reliable in most cases (Section~\ref{sec:discussion}).
\section{ADJUSTING COMPATIBILITY}
\label{sec:adjusting_comp}
\kg{can we define the setting, there is a machine learning model that generates hypothesis, there are features 
that include some aspects of user profiles, etc.} \ym{ Maybe the paragraph that comes right after Definition 2.1. explains the setting?  And I'm not sure how to include your second comment. I think we assume user histories have individual feature distributions, instead of assuming that there are specific features that describe the users. Should I say this? Maybe I misunderstood your comment}
\par Let $x$ be an instance in some data-set, $h_i$ be the prediction model of version $i$ of the AI-system such that $h_i(x)$ is the label predicted for instance $x$, and $h^*(x)$ be the correct label for that instance. In other words, $h_1$ is the model prior to the update to the AI-system (the ``pre-update" model) and $h_2$ is the model after the update (the ``post-update" model). Bansal et. al. \cite{bansal2019updates} propose a mathematical definition for the compatibility score of an update to an AI-system.
\begin{definition}{The \textit{compatibility score} of an update:}
\begin{equation}
\label{eq:compatibility}
C(h_1, h_2)=\frac{\sum_{x}\mathbbm{1}\big(h_1(x)=h^*(x)\big)\cdot\mathbbm{1}\big(h_2(x)=h^*(x)\big)}{\sum_{x}\mathbbm{1}\big(h_1(x)=h^*(x)\big)}
\end{equation}
\end{definition}
\noindent Where $\mathbbm{1}$ is the indicator function that returns $1$ if the expression received as the argument is $true$ and returns $0$ otherwise. The compatibility score of an update approaches 1 as the post-update model $h_2$ makes less new mistake and approaches 0 as the opposite occurs.
\begin{definition}{A \textit{new mistake} is a mistake introduced by the update, such that $h^*(x)=h_1(x)\neq h_2(x)$.}
\end{definition}
In order to increase the compatibility score of an update, the notion of the dissonance of an instance $x$ given an update to the system was introduced by Bansal et al \cite{bansal2019updates}.
\begin{definition}
The \textit{dissonance} of an instance given an update:
\begin{equation}
\label{eq:dissonance}
D(x,h_1,h_2)=\mathbbm{1}\big(h_1(x)=h^*(x)\big)\cdot L(x,h_2)
\end{equation}
\end{definition}
\kg{say that the dissonance is an updated loss function that considers compatability?} \ym{ I added ``is a loss function" in the second sentence of this paragraph, not sure if it's enough}
Where $L(x,h_i)$ is a regular loss function that penalizes a model $h_i$ for wrongly predicting the target label of an instance $x$ (e.g. cross-entropy loss). The dissonance function $D$ penalizes a low compatibility score by penalizing only new mistakes (as defined above), so by minimizing the loss incurred by it, the compatibility score of the update will increase as fewer new mistakes are introduced by it. The loss function to be used in the re-training (update) of an AI system proposed by Bansal et al. \cite{bansal2019updates} is a weighted sum of a regular loss $L$ and a loss incurred by the dissonance function $D$:
\begin{equation}
\label{eqn:non-personalized_loss}
L_c(x,h_1,h_2,\lambda)=(1-\lambda)\cdot L(x,h_2)+\lambda\cdot D(x,h_1,h_2)
\end{equation}
The trade-off between compatibility and accuracy of the post-update model $h_2$ can be adjusted by modifying the value of the dissonance weight $\lambda$. Increasing the value of $\lambda$ will likely increase the compatibility score of $h_2$ while simultaneously decreasing its accuracy as it's forced to make predictions similar to those the pre-update model $h_1$ makes.
\par Returning to the example with the doctor that deals mainly with face skin moles, once both the regular loss and the loss from dissonance are taken into consideration when updating the system (using Equation~\ref{eqn:non-personalized_loss}) fewer new mistakes will be introduced by the update, increasing the likelihood that the doctor will continue trusting the system's predictions at a certain expense of the system's overall accuracy.
\section{PERSONALIZATION OF UPDATES}
\label{sec:personalization}
\begin{figure}[t]
     \centering
     \includegraphics[width=0.45\textwidth]{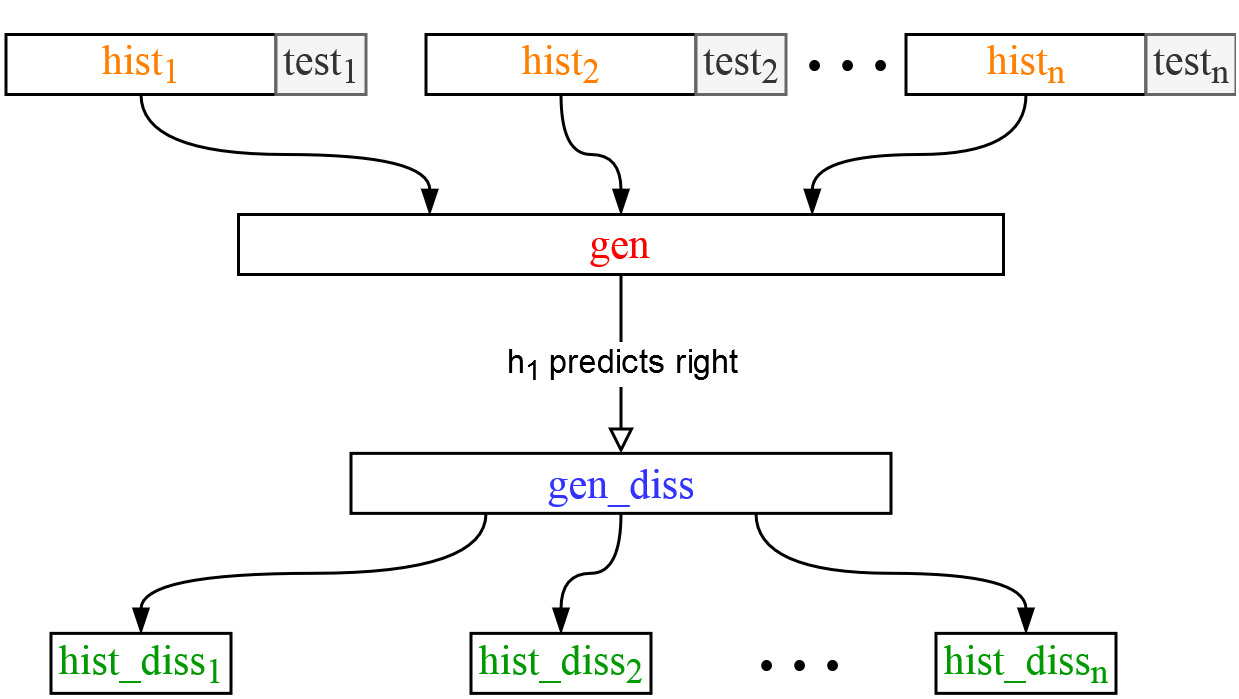}
     \captionsetup{width=.9\linewidth}
     \caption{The four types of subsets weighted by $W$ (Equation~\ref{eqn:sample_weights}). The subset $hist_i$ is the history of user $i$ and $test_i$ is the subset of $hist_i$ reserved for testing the trained models on that user. The subset $gen$ is the whole train set and, lastly, $gen\_ diss$ and $gen\_ diss$ are the subsets of $gen$ and $hist$ for which the pre-update model $h_1$ predicts the correct labels, respectively.}
    \label{fig:component_description}
\end{figure}
Our hypothesis is that extending Equation~\ref{eqn:non-personalized_loss} to include a notion of personalization towards the history of interaction of a particular user can yield a better compatibility-accuracy trade-off for that user. Let us re-frame Equation~\ref{eqn:non-personalized_loss} as an objective function that determines the relative weight of each instance in the train set:
\begin{equation}
\label{eqn:non-personalized_sample_weights}
(1-\lambda)\cdot gen+\lambda\cdot gen\_ diss
\end{equation}
Where $gen$ and $gen\_ diss$ (see description in Figure~\ref{fig:component_description}) are analogous to the regular loss $L$ and the dissonance loss $D$, respectively. To perform the personalization, we expand Equation~\ref{eqn:non-personalized_sample_weights} by adding two more components that relate only to instances that belong to the history of interaction of a specific user:
\begin{equation}
\label{eqn:sample_weights}
\begin{aligned}
&(1-\lambda)\cdot (w_{gen}\cdot gen+w_{hist}\cdot hist)+\\&\lambda\cdot (w_{gen\_ diss}\cdot gen\_ diss+w_{hist\_ diss}\cdot hist\_ diss)
\end{aligned}
\end{equation}
See Figure~\ref{fig:component_description} for a description of each one of these subsets. Each subset $i$ is weighted by a parameter $w_i$ and – as in Equation~\ref{eqn:non-personalized_loss} – by the dissonance weight $\lambda$ whose value will determine the resulting point in the compatibility-accuracy trade-off. This personalized objective function can be represented by a vector containing the component weights from Equation~\ref{eqn:sample_weights}:
\begin{equation}
\label{eqn:weight_vector}
W=(w_{gen},w_{gen\_ diss},w_{hist},w_{hist\_ diss})
\end{equation}
For instance, the model resulting from $W=(1,1,0,0)$ has the same objective function defined by Equation~\ref{eqn:non-personalized_loss}, where only the whole train set (represented by $gen$ or the $L$ from Equation~\ref{eqn:non-personalized_loss}) and its subset for which the pre-update model predicts the correct labels (represented by $gen\_ diss$ or the $D$ from Equation~\ref{eqn:non-personalized_loss}) are considered. We consider this to be our baseline – a model that isn't personalized towards any particular user. As another example, $W=(0,0,1,1)$ produces a model that considers only the train instances that belong to a specific user's history of interaction, ignoring the histories of any other users. Not all combinations of weights in $W$ produce models that yield a compatibility-accuracy trade-off, e.g., when $W=(1,0,1,0)$ the two dissonant components ($gen\_ diss$ and $hist\_ diss$) are ignored so varying the value of $\lambda$ doesn't affect the resulting predictions.
\section{METHODOLOGY}
\label{sec:methodology}
To support the hypothesis that the compatibility-accuracy trade-off with respect to a user can be improved by personalizing the objective function to the user's history, we conducted a series of experiments with various data-sets where we compared the performances of the various models by the following metric:
\begin{definition}
\label{def:AUTC}
The \textit{Area Under the Trade-off Curve} (AUTC) is the area under a compatibility-accuracy trade-off curve computed by the trapezoidal rule on the set of points that form it.
\end{definition}
\begin{table}
\centering
\caption{Models tested in the experiments. See Figure~\ref{fig:component_description} for an explanation of what each column in this table represents.}
\begin{tabular}[t]{lcccc}
\hline
Model name&$w_{gen}$&$w_{gen\_ diss}$&$w_{hist}$&$w_{hist\_ diss}$\\
\hline
baseline&1&1&0&0\\
L1&0&0&1&1\\
L2&0&1&1&0\\
L3&0&1&1&1\\
L4&1&0&0&1\\
L5&1&0&1&1\\
L6&1&1&0&1\\
L7&1&1&1&0\\
L8&1&1&1&1\\
\hline
\end{tabular}
\label{tab:models}
\end{table}
\begin{figure}[t]
     \centering
     \includegraphics[width=0.45\textwidth]{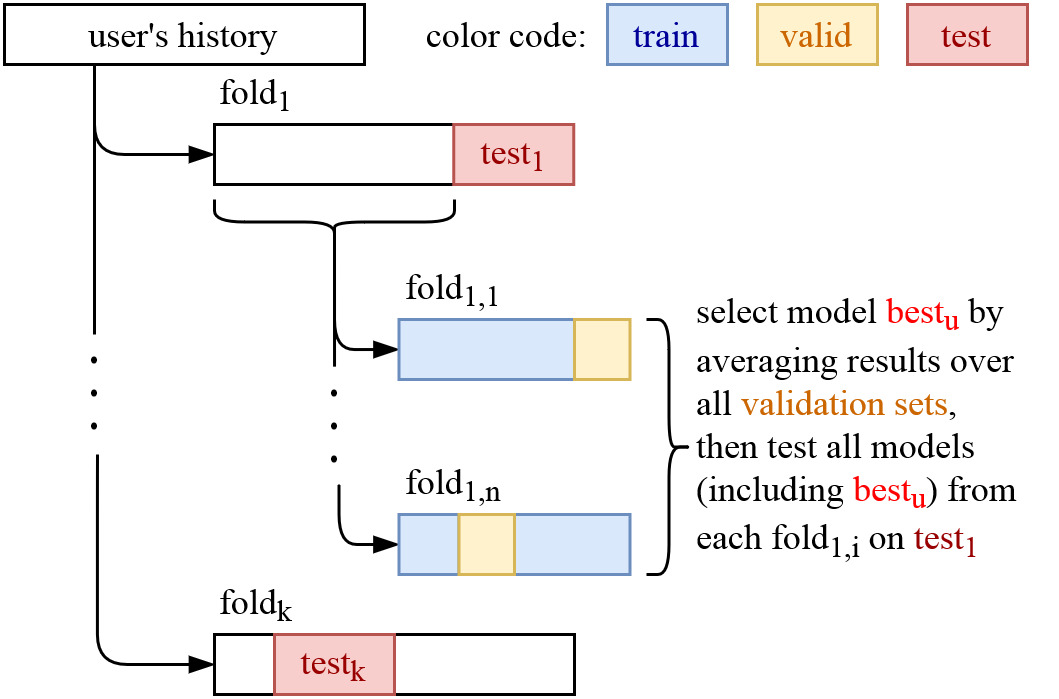}
     \captionsetup{width=.9\linewidth}
     \caption{ Organization of folds and inner folds employed to obtain the test results.}
    \label{fig:inner folds}
\end{figure}
\begin{figure*}
     \centering
     \includegraphics[width=0.9\textwidth]{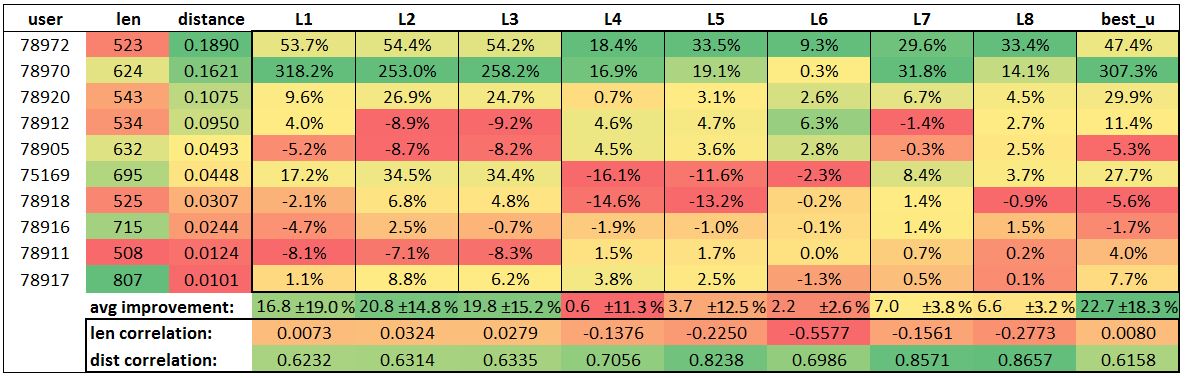}
     \captionsetup{width=.95\linewidth}
     \caption{ Results of the experiment with the ASSISTments data-set (see Section~\ref{sec:actual_users}). The ``len" and ``distance" columns indicate the length of the user's history the Wasserstein distance between its distribution and the distribution of the whole train set, respectively. The columns with the model names (see the definitions of these models in Table~\ref{tab:models}) indicate the percent improvement in AUTC relative to the baseline averaged over all folds in the experiment, along with one standard deviation. The ``len correlation" and ``dist correlation" rows indicate the correlation that the ``len" and ``distance" columns have with the values in the model columns.}
     \label{fig:assistment_correlations}
\end{figure*}
\noindent In each experiment we tested the models defined in Table~\ref{tab:models}, which correspond to all the possible vectors $W=(w_{gen},w_{gen\_ diss},w_{hist},w_{hist\_ diss})$ such that each $w_i\in W$ is either 0 or 1 and that produce a compatibility-accuracy trade-off curve (See last paragraph of Section~\ref{sec:personalization} for an example of $W$ that doesn't produce such a curve). This makes up a total of 9 models to train and test in each experiment, where the baseline $W=(1,1,0,0)$ is the model that gives the same importance to all instances regardless of which user history they belong to. Additionally, we determined which model produced the best trade-off (the one with the maximal AUTC) for each user $u$ by testing all models in $u$'s validation set and named this model $best_u$.
\par Here is the outline of each experiment, which is visually assisted by Figure~\ref{fig:inner folds}. Let $H_u$ be the set of instances that belong to the history of user $u$:
\begin{enumerate}[label*=\arabic*.]
\item For each fold$_i$ in a k-fold cross validation, shuffle\footnote{We performed the shuffling because our method needs the user's future interactions to resemble the current ones in order to work.} $H_u$ for all $u$ and then perform:
\begin{enumerate}[label*=\arabic*.]
\item Split $H_u$ for each $u$ into two sets – $test_u$ (for testing) and $H_u\setminus test_u$ (to be split multiple times into training and validation sets).
\item For each inner fold$_{i,j}$ in an n-fold cross validation performed inside\footnote{The inner folds' purpose is to increase the probability that the best model $best_u$ selected for $u$ using $u$'s validation set will be the best performing one in $u$'s test set as well.} fold$_i$, shuffle $H_u\setminus test_u$ and split it into a train set $train_u$ and a validation set $val_u$ and perform:
\begin{enumerate}[label*=\arabic*.]
\item Define the subset $gen\leftarrow \bigcup_u{train_u}$ and select a small subset from it (usually around 5\%) to train the pre-update model $h_1$ and use this model to generate $gen\_ diss$ accordingly (see Figure~\ref{fig:component_description}). Similarly, for each $u$ define $hist\leftarrow train_u$ and use $h_1$ to generate $hist\_ diss$ accordingly.
\item Train the various models (each model with a different set of subset weights $W$, see Section~\ref{sec:personalization}) using the subsets defined in the previous step and test them on $val_u$ and $test_u$ for each $u$.
\end{enumerate}
\item For each $u$, select the best model $best_u$ by averaging the trade-offs that resulted from testing all the models on the validation sets $val_u$ from each inner fold$_{i,j}$, and test all the models on $test_u$.
\end{enumerate}
\item Average the trade-offs over each test set in each inner fold$_{i,j}$ from each fold$_i$ to produce the final test plots (shown in Section~\ref{sec:results})    
\end{enumerate}
We employed sklearn's implementation of Regression Trees \cite{scikit-learn} as the predictive model in the experiments, for which we set the parameter ``sample\_weights" according to the subset weights vector $W$ that corresponds to the model to be tested. The implementation can be accessed via the following link: \url{https://github.com/jonmartz/CompatibleUpdates}.
%
% \newpage
\section{RESULTS}
\label{sec:results}
\par We conducted the experiment with various data-sets, some with information from real-world users for which the personalization was aimed, and some without them for which the personalization was aimed towards categories of some categorical feature of the dataset, e.g., if the selected column is ``gender" then there are two ``users" in the experiment – one ``male" and one ``female". This may be interesting since this is a way of personalizing towards groups of users. The description for all the following tables can be found in Figure~\ref{fig:assistment_correlations}. As a reminder, the baseline we compare the personalized models with corresponds to the subset weights $W=(1,1,0,0)$ where no personalization is performed (see Equation~\ref{eqn:weight_vector}). In every experiment, 10 folds each with 30 inner folds were performed (see Figure~\ref{fig:inner folds}) for increasing the statistical significance of the results and improving the chance that the model $best_u$ selected from the validation set of user $u$ is actually the best model in the test set of the same user.
\subsection{Individual users}
\label{sec:actual_users}
\par The results of the first experiment are shown in Figure~\ref{fig:assistment_correlations}, where the the data-set used is one of the ASSISTment data-sets\footnote{\url{https://sites.google.com/site/assistmentsdata/home/assistment-2009-2010-data}} \cite{wang2012student,pardos2010modeling,trivedi2011clustering,trivedi2010spectral}. In this classification task the goal is to predict whether students answer a question correctly on their first attempt. The train set's proportion was 80\% and the validation set's 10\%. For one of the users, $best_u$ managed to achieve over a 300\% improvement in AUTC relative to the baseline.
\par
\begin{figure*}
     \centering
     \includegraphics[width=0.9\textwidth]{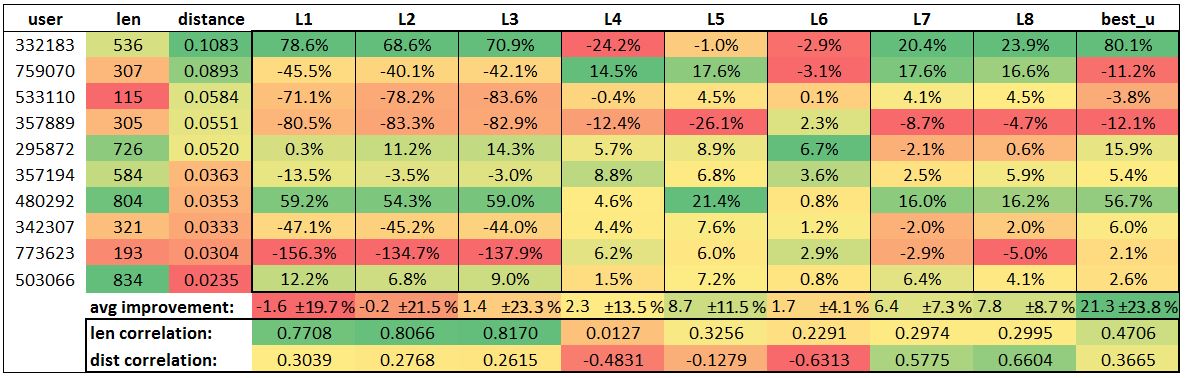}
     \captionsetup{width=.95\linewidth}
     \caption{ Results of the experiment with the edNet data-set. See Figure~\ref{fig:assistment_correlations} for description and Figure~\ref{fig:example_tradeoff} for visual reference.}
     \label{fig:ednet_correlations}
\end{figure*}

 The results of the second experiment are shown in Figure~\ref{fig:ednet_correlations}, where the data-set used is the recent edNet data-set\footnote{\url{https://github.com/riiid/ednet}} \cite{choi2019ednet}. In this classification task the goal is to predict if a student submits the correct answer for a question. The train set's proportion was 70\% and the validation set's 20\%. We extracted a large number of features for performing the task, e.g., amount of time spent on question, opportunities had for the skills involved in the question, etc., and also applied the pyBKT implementation\footnote{\url{https://github.com/CAHLR/pyBKT/blob/master/README.md}} of Bayesian Knowledge Tracing.
\subsection{Groups of users}
\label{sec:user-groups}
\par By ``groups of users" we refer to a grouping of instances other than by ``user id", e.g., by the ``gender" category of each instance or any other type of category. The improvement achieved by the personalized models in this section is smaller than the one achieved by them in the previous section. The reason may be that the feature distribution of groups of users is more similar to the feature distribution of the whole data-set ($hist$ and $gen$ from Figure~\ref{fig:component_description}, respectively) than that of individual users.
\par The results of the third experiment are shown in Figure~\ref{fig:adult_correlations}, where the data-set used is the well-known Adult data-set\footnote{\url{https://archive.ics.uci.edu/ml/datasets/Adult}}. In this classification task the goal is to predict whether a person earns more than 50K a year and the personalization was performed towards the categorical column ``relationship". The train set's proportion was 80\% and the validation set's 10\%.
\par The results of the fourth and last experiment are shown in Figure~\ref{fig:recidivism_correlations}, where the data-set used is the Recidivism data-set\footnote{\url{https://www.propublica.org/datastore/dataset/compas-recidivism-risk-score-data-and-analysis}}. In this classification task the goal is to predict if a criminal defendant will re-offend and the personalization was performed towards the categorical column ``race". The train set's proportion was 80\% and the validation set's 10\%. Here, the personalization provides only a small improvement relative to the baseline, but at least the results indicate that the mechanism we propose for selecting $best_u$ seems to be reliable.
\subsection{Discussion}
\label{sec:discussion}
The general trend than can be observed from the results is that most of the time at least some of the personalized models provide a better compatibility-accuracy trade-off (larger AUTC) than the one the baseline provides, but the set of models that achieves this varies between experiments. For instance, in most results the models $L_2$ and $L_3$ perform better than the baseline, but in the results from Figure~\ref{fig:ednet_correlations} $L_2$ performs even worse than the baseline and $L_3$ only slightly better.
\par However, in all four of the experiments, the model $best_u$ selected using each user's validation set was also the best performing one in average on the users' test set (in each fold separately, as explained in Section~\ref{sec:methodology}), and also had the smallest standard deviation relative to its average performance. This may be a solid indication that the mechanism we propose for selecting the best way of personalizing for each user works well.
\par
\begin{figure*}
     \centering
     \begin{subfigure}{0.9\textwidth}
     \includegraphics[width=0.9\textwidth]{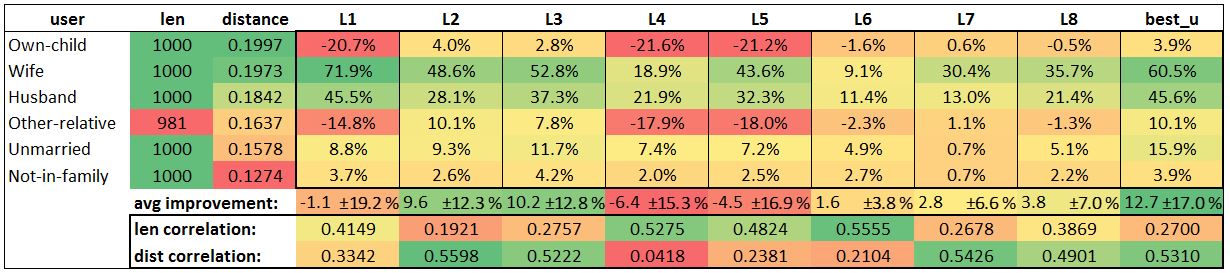}
     \caption{Adult data-set}
    \label{fig:adult_correlations}
     \end{subfigure}\hspace{0mm}
    \begin{subfigure}{0.9\textwidth}
     \includegraphics[width=0.9\textwidth]{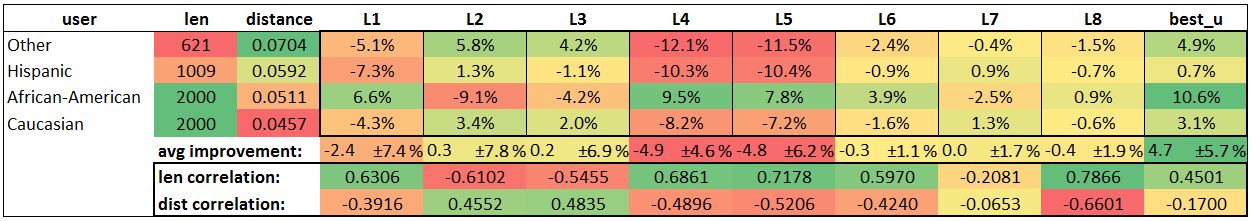}
     \caption{Recidivism data-set}
    \label{fig:recidivism_correlations}
     \end{subfigure}\hspace{0mm}
     \captionsetup{width=.95\linewidth}
     \caption{ Results of experiments from Section~\ref{sec:user-groups}. See description in Figure~\ref{fig:assistment_correlations}.}
     \label{fig:categorical_correlations}
\end{figure*}

 Is it always safe to assume that the best performing model on the training or validation set will be the best performing one on the test set as well? And what factors affect the performance of the personalization models relative to the baseline's? This depends on the degree to which the train and validation sets are representative of the test set, a property that is affected by multiple factors. We analyse how two of those factors correlate with the reliability of $best_u$ and the rest of the models:
\begin{itemize}
    \item The length of the validation set. We measure this factor by the length of the user's history since we are interested only in seeing how this correlates with the improvement relative to the baseline's performance in the same experiment, so we disregard what the proportion of the train set used for the validation was.
    \item The similarity between the feature distributions of the validation and test sets. We measure this factor as the Wasserstein distance \cite{vallender1974calculation} between those two distributions.
\end{itemize}
The measured Pearson correlation between these two factors and the percent improvement in AUTC relative to the baseline (shown in the last two rows of Figures \ref{fig:assistment_correlations} to \ref{fig:categorical_correlations}) varies greatly between the 4 classification tasks. For instance, in Figure~\ref{fig:assistment_correlations} for all models the history length shows almost no correlation while the Wasserstein distances show a large correlation, and in Figure~\ref{fig:ednet_correlations} the opposite is true for some of the models. The difference between these two experiments is that in the edNet experiment there were users with very short histories, which may explain the discrepancy – rather than being linearly correlated with this metric, the improvement provided by personalization depends on the user's history length being above a certain threshold (that varies depending on the classification task) for being capable of learning meaningful information about the user. The performance of $best_u$ was generally less correlated to these two factors, which means that $best_u$ can generally perform better than the baseline independently of these factors.
\par In general, the Wasserstein distance between the user's history and the whole data-set seems more correlated to the performance of the personalization models than the history length metric. We speculate that if the user's history is very different from the whole data-set set, giving a larger penalty for mistakes in instances that resemble the ones in this history will produce better trade-offs. And conversely, when the user's history is very similar to the whole train set, doing this will be less beneficial – or even detrimental – as the whole data-set could be considered an extension of this user's history.
\par A downside of the personalization method proposed in this paper is that it's not designed to handle significant changes in the way a user interacts with the system, since it assumes that the user's future interactions will resemble the current ones. In other words, the computations are atemporal.
\section{RELATED WORK}
\kg{can move earlier} \ym{ I thought putting it here to relate the models we proposed to other work after the experiments section so I don't have to mention models that haven't been presented yet. Do you think it's more adequate to put this section right after the introduction?}
This paper is strongly related to recent work of Bansal et al. \cite{bansal2019updates} that introduces the idea of the compatibility score of an update (Equation~\ref{eq:compatibility}) and proposes a method for increasing this score by employing a customized loss function (Equation~\ref{eqn:non-personalized_loss}) where an additional weighted penalty is given for new mistakes – mistakes that the pre-update version of the model didn't make. They showed that forcing the update to be more compatible generally decreases the accuracy of the updated model, generating a compatibility-accuracy trade-off. We expand this method by adding a notion of personalization towards each user (or any other type of subset of the train set) with the goal of producing better compatibility-accuracy trade-offs (i.e. with a larger AUTC, see Definition~\ref{def:AUTC}). 
\par The underlying idea behind the method proposed by Bansal et. al. \cite{bansal2019updates} (and therefore behind the method proposed here as well) is similar to several other works. One such example is model ensemble \cite{opitz1999popular}, AdaBoost \cite{freund1996experiments} in particular – in both that method and ours, an additional penalty is given for a certain type of mistakes that depends on a previously trained model. In AdaBoost this additional penalty is given for mistakes the previous model also made, and in our method (and Bansal et. al.'s) it is given for mistakes the previous model didn't make. A difference between these two methods is that, after the last model in the line is finished training, in AdaBoost all the previous models are kept and considered when making the final predictions and in our method all the previous models are discarded (in our case only one, the pre-update model). It could be interesting to explore the theoretical similarities between these two methods, since model ensemble enjoys a vast theoretical framework \cite{freund1996experiments}.
\par Choosing the best model for each user ($best_u$) is related to research on methods for choosing the best expert \cite{herbster1998tracking}, but in our work we simply calculate the AUTC of the models on a validation set to determine this. Further implementation of the ideas born in that research may improve the reliability of this selection. Several other works relate to the personalization of AI-models to users, but do not address the personalization of updates to these systems or any notion of dissonance between two versions of the system. For instance, for the ASSISTment data-set mentioned in previous sections, work was performed on individualizing student models  \cite{wang2012student,pardos2010modeling} and on clustering the students \cite{trivedi2011clustering,trivedi2010spectral} (related to the idea of grouping the train instances by some factor, see Section~\ref{sec:user-groups}) with the goal of improving the accuracy of the predictions.
\par Besides the work mentioned above, much work has been done in the field of human-AI interactions. The compatibility of an update to an AI-system is closely related to the \nth{14} Guideline for Human-AI Interactions from Amershi et al.'s work \cite{amershi2019guidelines} described as \emph{``Update and adapt cautiously: Limit disruptive changes when updating and adapting the
AI system’s behaviors''} i.e. making sure that the post-update version conforms to the user's mental model of the system that developed during the user's interaction with the pre-update version of the system. It is related also to the \nth{5} step in an article from Google Design \cite{lovejoy2017} that states the importance of making sure that the AI-system and the user's model evolve in tandem. For references to more related work on human-AI interaction and the field of AI-advised human decision making, refer to the related work section in Bansal et al.'s paper \cite{bansal2019updates}.
\section{CONCLUSION}
\par Update compatibility is very likely to be important for the adequate functioning of human-AI teams \cite{bansal2019beyond, bansal2019updates} (Section~\ref{sec:intro}). Previous work addressed the problem of making updates compatible by developing a loss function that delivers an increased penalty for mistakes that the pre-update one didn't make, and showed that a trade-off between the compatibility and accuracy of the update exists \cite{bansal2019updates} (Section~\ref{sec:adjusting_comp}). We extend this approach by adding the notion of personalization towards each user with the goal of producing better trade-offs (Section~\ref{sec:personalization}) and propose a framework for selecting the best way of performing this personalization (Section~\ref{sec:methodology}). We showed experimental results that indicate that considering this personalization can yield better trade-offs (and significantly better for some of the users) than a baseline model that doesn't consider it (Section~\ref{sec:results}) and analyzed some of the factors that affect the magnitude of this improvement (Section~\ref{sec:discussion}). This analysis indicated two things. First, that our method can be reliably employed given that the length of the user's history is longer than a certain threshold. And second, that as the distribution of this history deviates more from the distribution of the whole data-set, the bigger improvements we should expect from employing out method.
%
% \newpage
\bibliography{ecai}

\begin{thebibliography}{10}

\bibitem{amershi2019guidelines}
Saleema Amershi, Dan Weld, Mihaela Vorvoreanu, Adam Fourney, Besmira Nushi,
  Penny Collisson, Jina Suh, Shamsi Iqbal, Paul~N Bennett, Kori Inkpen, et~al.,
  `Guidelines for human-ai interaction', in {\em Proceedings of the 2019 CHI
  Conference on Human Factors in Computing Systems}, pp. 1--13, (2019).

\bibitem{bansal2019beyond}
Gagan Bansal, Besmira Nushi, Ece Kamar, Walter~S Lasecki, Daniel~S Weld, and
  Eric Horvitz, `Beyond accuracy: The role of mental models in human-ai team
  performance', in {\em Proceedings of the AAAI Conference on Human Computation
  and Crowdsourcing}, volume~7, pp. 2--11, (2019).

\bibitem{bansal2019updates}
Gagan Bansal, Besmira Nushi, Ece Kamar, Daniel~S Weld, Walter~S Lasecki, and
  Eric Horvitz, `Updates in human-ai teams: Understanding and addressing the
  performance/compatibility tradeoff', in {\em Proceedings of the AAAI
  Conference on Artificial Intelligence}, volume~33, pp. 2429--2437, (2019).

\bibitem{choi2019ednet}
Youngduck Choi, Youngnam Lee, Dongmin Shin, Junghyun Cho, Seoyon Park, Seewoo
  Lee, Jineon Baek, Byungsoo Kim, and Youngjun Jang, `Ednet: A large-scale
  hierarchical dataset in education', {\em arXiv preprint arXiv:1912.03072},
  (2019).

\bibitem{freund1996experiments}
Yoav Freund, Robert~E Schapire, et~al., `Experiments with a new boosting
  algorithm', in {\em icml}, volume~96, pp. 148--156. Citeseer, (1996).

\bibitem{herbster1998tracking}
Mark Herbster and Manfred~K Warmuth, `Tracking the best expert', {\em Machine
  learning}, {\bf 32}(2),  151--178, (1998).

\bibitem{lovejoy2017}
Holbrook Lovejoy.
\newblock Human-centered machine learning: 7 steps to stay focused on the user
  when designing with ml.
\newblock
  \url{https://medium.com/google-design/human-centered-machine-learning-a770d10562cd},
  2017.
\newblock Accessed Feb 2020.

\bibitem{opitz1999popular}
David Opitz and Richard Maclin, `Popular ensemble methods: An empirical study',
  {\em Journal of artificial intelligence research}, {\bf 11},  169--198,
  (1999).

\bibitem{pardos2010modeling}
Zachary~A Pardos and Neil~T Heffernan, `Modeling individualization in a
  bayesian networks implementation of knowledge tracing', in {\em International
  Conference on User Modeling, Adaptation, and Personalization}, pp. 255--266.
  Springer, (2010).

\bibitem{scikit-learn}
F.~Pedregosa, G.~Varoquaux, A.~Gramfort, V.~Michel, B.~Thirion, O.~Grisel,
  M.~Blondel, P.~Prettenhofer, R.~Weiss, V.~Dubourg, J.~Vanderplas, A.~Passos,
  D.~Cournapeau, M.~Brucher, M.~Perrot, and E.~Duchesnay, `Scikit-learn:
  Machine learning in {P}ython', {\em Journal of Machine Learning Research},
  {\bf 12},  2825--2830, (2011).

\bibitem{sahoo2019deepreco}
Abhaya~Kumar Sahoo, Chittaranjan Pradhan, Rabindra~Kumar Barik, and
  Harishchandra Dubey, `Deepreco: deep learning based health recommender system
  using collaborative filtering', {\em Computation}, {\bf 7}(2), ~25, (2019).

\bibitem{trivedi2010spectral}
Shubhendu Trivedi, Zachary Pardos, G{\'a}bor S{\'a}rk{\"o}zy, and Neil
  Heffernan, `Spectral clustering in educational data mining', in {\em
  Educational Data Mining 2011}, (2010).

\bibitem{trivedi2011clustering}
Shubhendu Trivedi, Zachary~A Pardos, and Neil~T Heffernan, `Clustering students
  to generate an ensemble to improve standard test score predictions', in {\em
  International conference on artificial intelligence in education}, pp.
  377--384. Springer, (2011).

\bibitem{vallender1974calculation}
SS~Vallender, `Calculation of the wasserstein distance between probability
  distributions on the line', {\em Theory of Probability \& Its Applications},
  {\bf 18}(4),  784--786, (1974).

\bibitem{wang2012student}
Yutao Wang and Neil~T Heffernan, `The student skill model', in {\em
  International Conference on Intelligent Tutoring Systems}, pp. 399--404.
  Springer, (2012).

\end{thebibliography}
\end{document}